\documentclass{sig-alternate}

\usepackage[ruled,vlined]{algorithm2e}
\usepackage{amsmath,bm}
\usepackage{graphicx}
\usepackage{epstopdf}
\usepackage{epsfig}

\usepackage{tabularx}
\usepackage{subfig}

\usepackage{tikz}
\usetikzlibrary{positioning}
\usetikzlibrary{shapes,snakes}

\newtheorem{theorem}{Theorem}

\allowdisplaybreaks

\begin{document}
%
\conferenceinfo{KDD}{'15 Sydney Australia}

\title{Infinite Author Topic Model based on Mixed Gamma-Negative Binomial Process}

\numberofauthors{5}
\author{
\alignauthor
Junyu Xuan\\
       \affaddr{University of Technology Sydney}\\
       \affaddr{15 Broadway}\\
       \affaddr{Sydney, Australia}\\
       \email{Junyu.Xuan@student.uts.edu.au}
\alignauthor
Jie Lu\\
       \affaddr{University of Technology Sydney}\\
       \affaddr{15 Broadway}\\
       \affaddr{Sydney, Australia}\\
       \email{Jie.Lu@uts.edu.au}
\alignauthor
Guangquan Zhang\\
       \affaddr{University of Technology Sydney}\\
       \affaddr{15 Broadway}\\
       \affaddr{Sydney, Australia}\\
       \email{Guangquan.Zhang@uts.edu.au}
\and
\alignauthor
Richard Yi Da Xu\\
       \affaddr{University of Technology Sydney}\\
       \affaddr{15 Broadway}\\
       \affaddr{Sydney, Australia}\\
       \email{Yida.Xu@uts.edu.au}
\alignauthor
Xiangfeng Luo\\
       \affaddr{Shanghai University}\\
       \affaddr{99 Shangda Road}\\
       \affaddr{Shanghai, China}\\
       \email{luoxf@shu.edu.cn}
}

\maketitle
\begin{abstract}

Incorporating the side information of text corpus, i.e., authors, time stamps, and emotional tags, into the traditional text mining models has gained significant interests in the area of information retrieval, statistical natural language processing, and machine learning. One branch of these works is the so-called Author Topic Model (ATM), which incorporates the authors's interests as side information into the classical topic model. However, the existing ATM needs to predefine the number of topics, which is difficult and inappropriate in many real-world settings. In this paper, we propose an Infinite Author Topic (IAT) model to resolve this issue. Instead of assigning a discrete probability on fixed number of topics, we use a stochastic process to determine the number of topics from the data itself. To be specific, we extend a gamma-negative binomial process to three levels in order to capture the author-document-keyword hierarchical structure. Furthermore, each document is assigned a mixed gamma process that accounts for the multi-author's contribution towards this document. An efficient Gibbs sampling inference algorithm with each conditional distribution being closed-form is developed for the IAT model. Experiments on several real-world datasets show the capabilities of our IAT model to learn the hidden topics, authors' interests on these topics and the number of topics simultaneously.

\end{abstract}



\section{Introduction}

Traditional text mining algorithms only model the text corpus with two levels: document-word. Topic models are commonly regarded as the efficient tools for the text mining by learning the hidden topics \cite{2020481}. Recently, interests have been paid on the side information of the text corpus, which includes the conferences of the papers \cite{conference}, time stamps \cite{tot}, authors \cite{atm1,atm2}, entities \cite{entry}, emotion tags \cite{emotion} and other labels \cite{labels}. The incorporation of these side information into the classical topic models benefits a lot of real-world tasks. Among them, Author Topic Model (ATM) \cite{atm1,atm2,atm3} is proposed by adding a set of variables to the original topic model aiming to indicate and inference the interests of authors together with the hidden topics.

The ability to jointly learn the hidden topics and authors' interests on these topics has a variety of application scenarios. For example, 1) an academic recommendation system can recommend authors and/or papers with similar research interests to that of the input author; 2) detecting the most and least surprising papers for an author \cite{atm2}; 3) in an author-topic-based paper browser, a set of papers can be ranked according to authors and topics; 4) authors disambiguation \cite{Zhao2013}.

One drawback of the existing author topic model is that the number of hidden topics needs to be fixed in advance. This number is normally chosen with domain knowledge. By fixing the number of topics, ATM can then adopt Dirichlet and Multinomial distributions with the pre-defined dimension. However, limiting each document to have exactly fixed number of topics is apparently unrealistic for many real-world applications. In this paper, we propose an infinite author topic (IAT) model to relax this assumption. Instead of using fixed-dimensional distributions, stochastic processes are used: to be specific, the gamma-negative binomial process \cite{6636308} is extended to three levels for capturing the hierarchical structure: author-document-keyword. In this model, each document is assigned with a gamma process to express the interest of this document on the hidden topics instead of a vector with a fixed dimension. This gamma process can be simply considered as an infinite discrete distribution, and is parameterized by a base measure (another gamma process) that denotes the interest of the author of this document on the hidden topics.
However, a document normally has multiple authors, so we assign a document a mixed gamma process that is based on all the gamma processes of the authors of this document. Furthermore, introducing mixed gamma process will lead to intricacies in terms of model inference. Therefore, an efficient Gibbs sampling with closed-form conditional distributions is developed for the proposed model.
Experiments on the two real-world datasets show the capability of our model to learn both the hidden topics and the number of topics, simultaneously.

The main contributions of this paper are,
\begin{enumerate}
  \item propose a new nonparametric Bayesian model to relax the fixed topic number assumption of the traditional author topic models;
  \item design an efficient Gibbs sampling inference algorithm for getting the solution of the proposed model.
\end{enumerate}

The rest paper is structured as follows. Section 2 briefly introduces the related work. Section 3 describes some preliminary knowledge. The IAT model is proposed and presented in Section 4 with its Gibbs sampling inference algorithm. Section 5 describes the IAT model experimental results using real-world datasets. Finally, Section 6 concludes this study with a discussion on future directions.

\section{Related Work}

In this section, we briefly review the related work of this study. The first part is about the topic models, and the second part is about nonparametric Bayesian learning.

\subsection{Topic Models}

Topic models \cite{blei2003latent} are Bayesian models with fixed-dimensional probability distributions. They are originally designed for the text mining task, which aim to discover the hidden topics in the text corpus to assist document clustering or classification. Due to their good extendibility and powerful representation, they have been successfully applied to many research areas, including analysis in image \cite{tmimage}, video \cite{tmvideo}, genetics \cite{tmpgeneric} and music \cite{tmmusic}. Among these extensions, author topic models \cite{atm1,atm2,atm3} were proposed to infer the hidden topics and author interests. The documents are supposed to be generated by its authors according to their interests over the hidden topics. This model will be explained with more details in Section 3.

ATM has attracted a lot of attentions from researchers working in the text mining area, because it provides an elegant way to incorporate the side (in this case, author) information of the documents for topic learning. This model can be extend to incorporate other side information of text corpus, such as emotional tags \cite{emotion}, conferences\cite{conference} and time stamps \cite{tot}.

\subsection{Nonparametric Bayesian Learning}

Nonparametric Bayesian learning is a key approach for learning the number of mixtures in a mixture model (also called model selection problem). Without predefining the number of mixtures, this number is supposed to be inferred from the data, i.e., let the data speak.

The idea of nonparametric Bayesian learning is to use the stochastic processes to replace the traditional fixed-dimensional probability distributions, such as Multinomial, Poisson, and Dirichlet. In order to avoid the limitation associated with fixed dimensions, Multinomial Process (MP), Poisson Process (PP) \cite{pp} and Dirichlet Process (DP) \cite{teh2010dirichlet} are used to replace former distributions because of their infinite properties.

The merit of these stochastic processes is that they let the data to determine the number of factors (in text mining task, topics). DP is a good alternative for the models with Dirichlet distribution as the prior. Many probabilistic models with fixed dimensions have been extended to the infinite ones by the help of stochastic processes: Gaussian Mixture Model (GMM) is extended to Infinite Gaussian Mixture Model (IGMM) \cite{rasmussen1999infinite} using DP; Hidden Markov Model is extended with infinite number of hidden states using Hierarchial Dirichlet Process \cite{teh2006hierarchical,Fox:AOAS2011}. Through the posterior inference (i.e., Markov chain Monte Carlo (MCMC) \cite{neal2000markov}), the number of the mixtures can be inferred. Other popular processes include beta process, gamma process, poisson process, multinomial process, negative binomial process (NBP) \cite{6636308,6802382} have also been used in the machine learning communities recently.

To summarize, nonparametric Bayesian learning \cite{Buntine} has been successfully used to extend many finite models and applied to many real-world applications. However, to the best of our knowledge, there has not been any works proposed to use NBP for author topic modelling. This paper addresses this shortcoming by proposing a mixed gamma negative binomial process to extend the finite author topic model to the infinite one.

\section{Preliminary Knowledge}

\begin{figure}[!t]
\centering
    \tikzstyle{rv}=[circle,
                        thick,
                        minimum size=1.2cm,
                        draw=black!80,
                        ]
    \tikzstyle{line}=[->,
                        solid,
                        line width=1pt,
                        draw=black!80,
                        ]
 \tikzstyle{dt}=[circle,
                        thick,
                        minimum size=1.2cm,
                        draw=black!80,
                        fill=gray
                        ]
    \begin{tikzpicture}[font=\Large,scale=0.75, rotate=270]
     	

    	\node[dt] (1) at (2, 0) {$a_d$};
        \node[rv] (2) at (2, -2) {$x_{d,n}$};
    	\node[rv] (3) at (2, -4) {$z_{d,n}$};
        \node[dt] (4) at (2, -6) {$w_{d,n}$};

    	\draw[thick, rounded corners] (0.5,-8.2) rectangle (3.5,1);
        \draw[thick, rounded corners] (3,-7.5) rectangle (1,-1);

    	\node[rv] (5) at (-1.5, 0) {$\alpha$};
        \node[rv] (6) at (-1.5, -2) {$\rho_a$};
    	\node[rv] (7) at (-1.5, -4) {$\phi_k$};
        \node[rv] (8) at (-1.5, -6) {$\beta$};

        \draw[thick, rounded corners] (-3,-1) rectangle (-0.5,-2.9);
        \draw[thick, rounded corners] (-3,-5) rectangle (-0.5,-3.1);

        \node at (-2.5,-3) [right] {\large A};
        \node at (-2.5,-5) [right] {\large K};

        \node at (2.4,-7.5) [right] {\large N};
        \node at (2.9,-8.2) [right] {\large D};

    	\path[line]
    		
            (1)		edge (2)
    		(2)		edge (3)
            (3)		edge (4)

    		(5)		edge (6)
            (6)		edge (3)
    		(7)		edge (4)
            (8)		edge (7)
    		
            ;

    \end{tikzpicture}
\caption{Classical Author-Topic Model}
  \label{graphatm}
\end{figure}
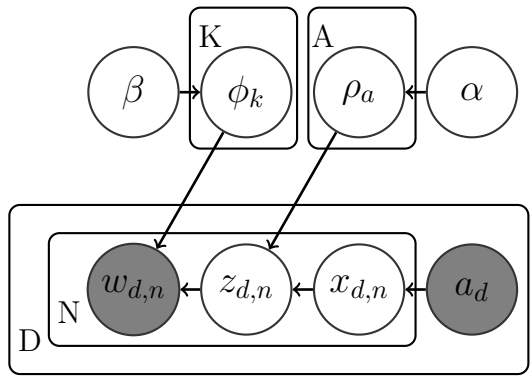

This section briefly introduces the related models which will be used in the rest of sections.

\subsection{Author Topic Model}

The Author Topic Model \cite{atm1,atm2,atm3} aims to learn the hidden topics from the papers and more importantly learn the authors' interests on these topics. Based on the classical LDA \cite{blei2003latent}, a set of new variables are introduced to indicate the authors' interests. The graphical representation of the model is shown in Fig. \ref{graphatm}, and the generative procedure is as follows,
\begin{equation}\label{atmequ}
\begin{aligned}
\rho_a &\overset{i.i.d}{\sim} Dirichlet(\alpha)\\
\phi_k &\overset{i.i.d}{\sim} Dirichlet(\beta)\\
x_{d,n} &\sim Unif(a_d)\\
z_{d,n} &\sim Discrete(\rho_{x_{d,n}})\\
w_{d,n} &\sim Discrete(\phi_{z_{d,n}})\\
\end{aligned}
\end{equation}
where $\{\rho_a\}_{a=1}^A$ denote the authors' interests on the topics and $a_d$ denotes the authors of a document. We can see from the Eq.(\ref{atmequ}) that the ATM is constructed by the fixed-dimensional probability distributions. One issue of this model is that the number of topics needs to be predefined, because the dimensions of the probability distributions need to be predefined. However, it is very difficult and not appropriate to predefine the topic number in many real-world scenarios.

\subsection{Gamma Negative Binomial Process}

\subsubsection{Gamma Process}

A gamma process $GaP(\alpha, H)$ \cite{gp2014} is a stochastic process, where $H$ is a base (shape) measure and $\alpha$ is the concentration (scale) parameter. It also corresponds to a complete random measure. Let $\Gamma = \{ (\pi_i, \theta_i)\}_{i=1}^{\infty}$ be a random realization of a Gamma process in the product space ${R}^+ \times \Theta$. Then, we have
\begin{equation}
\begin{aligned}
\Gamma &\sim GaP(\alpha, H) \\
&= \sum_{i=1}^{\infty} \pi_i \delta_{\theta_i}
\end{aligned}
\end{equation}
where $\delta(\cdot)$ is an indicator function, $\pi_i$ satisfies an improper gamma distribution $gamma(0, \alpha)$, and $\theta_i \sim H$. After the normalization of the $\{\pi\}$, we can get the famous Dirichlet process \cite{teh2010dirichlet}.

\subsubsection{Negative Binomial Process}

A negative binomial process $NBP(p, G_0)$ \cite{6636308} is also a stochastic process parameterized by a base measure $G_0$ and $p$. Similar with the gamma process, a realization of negative binomial process $X = \{(n_i, \theta_i)\}_{i=1}^{\infty}$ is also a set of points in product space $Z^+ \times \Theta$. Then, we have
\begin{equation}\label{nbpq}
\begin{aligned}
X &\sim NBP(p, G_0) \\
&= \sum_{i=1}^{\infty} n_i \delta_{\theta_i}
\end{aligned}
\end{equation}
where $\{n_i\}$ are integers, so negative binomial process is normally used for the counting model \cite{6802382}. Compared with Poisson process which is also suitable for the counting model, negative binomial process has a better variance-to-mean ratio (VMR) and the overdispersion level \cite{6636308}.

\subsubsection{Gamma-Negative Binomial Process}

\begin{figure}[!th]
\centering
    \tikzstyle{rv}=[circle,
                        thick,
                        minimum size=1.2cm,
                        draw=black!80,
                        ]
    \tikzstyle{line}=[->,
                        solid,
                        line width=1pt,
                        draw=black!80,
                        ]
    \tikzstyle{dt}=[circle,
                        thick,
                        minimum size=1.2cm,
                        draw=black!80,
                        fill=gray
                        ]
    \begin{tikzpicture}[font=\Large,scale=0.7]
     	

    	\node[rv] (1) at (0, 0) {$H$};
        \node[rv] (2) at (0, -3) {$\Gamma$};
    	\node[dt] (3) at (0, -6) {$X_d$};

        \node[rv] (5) at (2, 0) {$\alpha$};
        \node[rv] (6) at (2, -6) {$p_d$};

    	\draw[thick, rounded corners] (-1,-4.5) rectangle (3,-8);

    	\node[rv] (7) at (5, 0) {$H$};
        \node[rv] (8) at (5, -3) {$\Gamma$};
    	\node[rv] (9) at (5, -6) {$\Gamma_d$};
        \node[dt] (10) at (5, -9) {$X_d$};

        \node[rv] (11) at (7, 0) {$\alpha$};
        \node[rv] (12) at (7, -6) {$p_d$};

        \draw[thick, rounded corners] (4,-4.5) rectangle (8,-10);

        \node at (2.1,-7.5) [right] {\large D};
        \node at (7.2,-9.2) [right] {\large D};

    	\path[line]
    		
            (1)		edge (2)
    		(2)		edge (3)
            (5)		edge (2)
            (6)		edge (3)

            (7)		edge (8)
    		(8)		edge (9)
            (9)		edge (10)
            (11)		edge (8)
            (12)		edge (9)
    		    		
            ;

    \end{tikzpicture}
\caption{Gamma-Negative binomial process topic model}
  \label{graphgnb}
\end{figure}
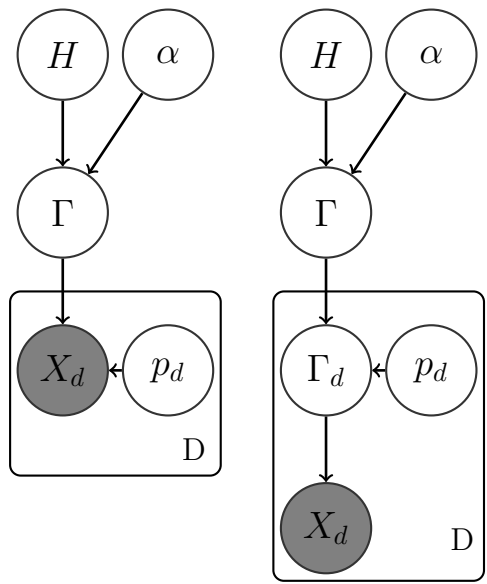

Normally, negative binomial process is used as the likelihood part of a Bayesian model. Like a negative binomial distribution $x \sim NB(r, p)$ which has two parameters: $r > 0$ and $p \in [0, 1]$, there are two kinds of priors for a negative binomial process: one is Gamma process \cite{6636308} as shown in Eq. (\ref{nbpq}); the other is the Beta process \cite{6802382} as $X \sim NBP(B, r)$. In this paper, we use the Gamma process prior. A gamma-negative binomial process-based topic model is proposed in \cite{6636308} as shown in Fig. \ref{graphgnb} and it can be represented as,
\begin{equation}
\begin{aligned}
\Gamma &\sim GaP(c_0, H)\\
X_d &\sim NBP(p_d, \Gamma)\\
\end{aligned}
\label{gnbtm1}
\end{equation}
where the base measure of the negative binomial process $\Gamma$ is a random measure from a gamma process. $X_d$ is for each document, and this hierarchial form makes the documents share a same base measure $\Gamma$.
This gamma-negative binomial process can be equivalently augmented as gamma-gamma-poisson process,
\begin{equation}
\begin{aligned}
\Gamma &\sim GaP(c_0, H)\\
\Gamma_d &\sim GaP\left(\frac{1-p_d}{p_d}, \Gamma \right)\\
X_d &\sim PP(\Gamma_d)\\
\end{aligned}
\label{gnbtm2}
\end{equation}
where $X_d \sim PP(\Gamma_d)$ is a Poisson process with parameter $\Gamma_d$. This augmentation, which is useful for the close-form model inference algorithm design, is equal to gamma-negative binomial process model in distribution.

\begin{table}[tb]
\caption{Notations used in this paper}\label{notations}
\centering
\begin{tabular}{c|l}
\hline
Notation               & description   \\
\hline
$D$                 & number of documents      \\
\hline
$A$            &  number of authors      \\
\hline
$N$            &  number of words      \\
\hline
$AD$            & author-document mapping matrix     \\
\hline
$DN$            & document-word mapping matrix     \\
\hline
$A_d$            & number of authors of document $d$   \\
\hline
\end{tabular}
\end{table}

In this paper, we will build an infinite author topic model based on this gamma-negative binomial process model.

\section{Infinite Author Topic Model}

In this section, we first propose our infinite author topic (IAT) model, and then introduce its Gibbs sampling strategy to inference the proposed model.

\subsection{Model Description}

\begin{figure}[!t]
\centering
    \tikzstyle{rv}=[circle,
                        thick,
                        minimum size=1.2cm,
                        draw=black!80,
                        ]
    \tikzstyle{line}=[->,
                        solid,
                        line width=1pt,
                        draw=black!80,
                        ]
    \tikzstyle{dt}=[circle,
                        thick,
                        minimum size=1.2cm,
                        draw=black!80,
                        fill=gray
                        ]
    \begin{tikzpicture}[font=\Large,scale=0.7]
     	

        \node[rv] (0) at (0, 3) {$H$};
    	\node[rv] (1) at (0, 0) {$\Gamma_0$};
        \node[rv] (2) at (0, -3) {$\Gamma_a$};
    	\node[rv] (3) at (0, -6) {$\Gamma_d$};
        \node[dt] (4) at (0, -9) {$X_d$};

        \node[rv] (5) at (2, 3) {$c_0$};
        \node[rv] (7) at (2, -3) {$c_a$};
        \node[rv] (6) at (2, -6) {$p_d$};

    	\draw[thick, rounded corners] (-1.3,-1.5) rectangle (3.3,-11);
        \draw[thick, rounded corners] (-1,-4.5) rectangle (3,-10);

        \node at (2.1,-9.5) [right] {\large D};
        \node at (2.1,-10.5) [right] {\large A};


        \node[rv] (10) at (5, 3) {$H$};
    	\node[rv] (11) at (5, 0) {$\Gamma_0$};

    	\node[rv] (13) at (5, -8) {$p_d$};
        \node[dt] (14) at (7, -10) {$X_d$};

        \node[rv] (15) at (7, 3) {$c_0$};
        \node[rv] (12) at (4.5, -3) {$\Gamma_{a_1}$};
        \node[rv] (17) at (6.5, -3) {$\Gamma_{a_2}$};
        \node at (7.5, -3) [right] {$\cdots$};
        \node[rv] (18) at (9.5, -3) {$\Gamma_{a_A}$};
        \node[rv] (16) at (7, -8) {$\Gamma_d$};

        \node[rv] (19) at (9.5, -5.35) {$c_a$};

    	\draw[thick, rounded corners] (3.5,-1.5) rectangle (10.5,-6.5);

        \draw[thick, rounded corners] (4,-6.8) rectangle (8,-11);


        \node at (4.0,-10.5) [right] {\large D};

        \node at (4.0,-6) [right] {\large A};

    	\path[line]
    		(0)		edge (1)
            (1)		edge (2)
    		(2)		edge (3)
            (5)		edge (1)
            (6)		edge (3)
            (3)		edge (4)
            (7)		edge (2)

            (10)		edge (11)
            (15)		edge (11)

            (11)		edge (12)
            (11)		edge (17)
            (11)		edge (18)

            (16)		edge (14)

            (13)		edge (16)
    		
            (12)		edge (16)
            (17)		edge (16)
            (18)		edge (16)

            (19)		edge (12)
            (19)		edge (17)
            (19)		edge (18)

            ;
    \end{tikzpicture}
    \caption{Gamma-Gamma-Negative Binomial Process Model (3GNB) (left one) and Infinite Author Topic Model (IAT) (right one)}
    \label{agnbtms}
\end{figure}
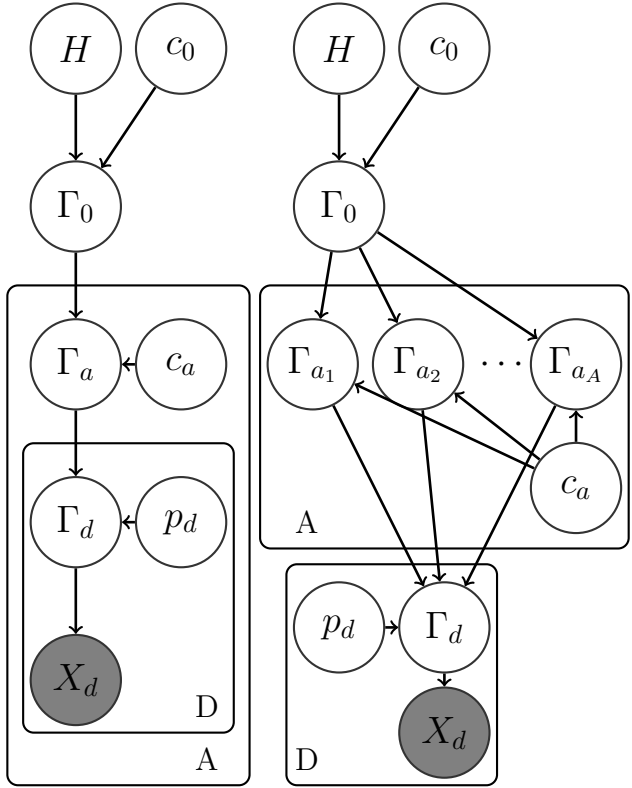

Consider the gamma-negative binomial process topic model in Eqs. (\ref{gnbtm1}) and (\ref{gnbtm2}) again: despite its successful, this model however is fundamentally the same as the basic topic models, which are used for modeling the data of two level hierarchy: document-keyword. Our aim is to extend topic model into three-level hierarchy: author-document-keyword.
So we add another gamma process level to capture the additional (author) level based on the gamma-negative binomial process topic model in Eq.(\ref{gnbtm2}) analogues to the hierarchical form of Hieratical Dirichlet Process \cite{teh2006hierarchical},
\begin{equation}
\begin{aligned}
\Gamma_0    &\sim GaP(c_0, H)\\
\Gamma_a    &\sim GaP(c_a, \Gamma_0)\\
\Gamma_d    &\sim GaP((1-p_d)/p_d, \Gamma^d_a)\\
X_d         &\sim PP(\Gamma_d)\\
\end{aligned}
\label{agnbtm}
\end{equation}
where $\Gamma_a$ is the new added level for the authors. We call this model three-level gamma-negative binomial process topic model (3GNB), which is graphically shown in the left subfigure of Fig. \ref{agnbtms}. However, there is a problem in 3GNB that it requires each document with only one author.

In the 3GNB model, each document is assigned a realization of gamma process,
\begin{equation}
\begin{aligned}
\Gamma_d &= \sum_{k=1}^{\infty} \pi_{d,k} \delta_{\theta_k}
\end{aligned}
\end{equation}
where $\theta_k$ denotes the $k$th topic and $\pi_{d,k}$ is the weight of $k$th topic. $\{\pi_{d,k}\}_{k=1}^{\infty}$ can be viewed as the interest of document $d$ on the topics. The number of topics can potentially be infinite and therefore justifies the infinity in the summation. However, since the data is limited, the learned topics will be also limited. Similar to the document, each author is also assigned a realization of gamma process,
\begin{equation}
\begin{aligned}
\Gamma_a &= \sum_{k=1}^{\infty} \pi_{a,k} \delta_{\theta_k}
\end{aligned}
\end{equation}
where $\{\pi_{a,k}\}_{k=1}^{\infty}$ is the weight of interests of author $a$ on the topics. In the 3GNB model, the base measure for a $\Gamma_d$ is from its author $\Gamma_a$. It can be seen as the `interest inheritance'.

In order to model in the setting where a document is with multiple authors, we combine all the gamma processes of every authors of a document together by
\begin{equation}
\begin{aligned}
\Gamma^d_a = \Gamma_{a_1} \oplus \Gamma_{a_2} \oplus \cdots \oplus \Gamma_{a_{A_d}}
\end{aligned}
\end{equation}
where $A_d$ is the number of authors of document $d$, $\oplus$ is the convex combination (each gamma process is with same weight in this paper) and $\Gamma^d_a$ is the mixed prior for $\Gamma_d$. We can see the mixed gamma process $\Gamma^d_a$ as the mixed interests of all the authors of a document. Then, the revised model is as follow
\begin{equation*}
\begin{aligned}
\Gamma_0 &\sim GaP(c_0, H)\\
\Gamma_a &\sim GaP(c_a, \Gamma_0)\\
\Gamma^d_a &= \Gamma_{a_1} \oplus \Gamma_{a_2} \oplus \cdots \oplus \Gamma_{a_{A_d}} \\
\Gamma_d &\sim GaP((1-p_d)/p_d, \Gamma^d_a)\\
X_d &\sim PP(\Gamma_d)\\
\end{aligned}
\end{equation*}
and the graphical representation is shown in Fig. \ref{agnbtms}. Some frequently used notations are explained in Table \ref{notations}.

\subsection{Model Inference}

It is difficult to perform posterior inference under infinite mixtures, a common work-around solution in nonparametric Bayesian learning is to use a truncation method. Truncation method is widely accepted, which uses a relatively big $K$ as the (potential) maximum number of topics. Under the truncation, the model can be expressed below as a good approximation to the infinite model,
\begin{equation*}\label{model0}
\begin{aligned}
\gamma_0                    &\sim Gamma(e_0, 1/f_0)\\
r_{0,k} | \gamma_0, c_0     &\sim Gamma(\gamma_0/K, 1/c_0)\\
r_{a,k} | r_0, c_a          &\sim Gamma(r_{0,k}, 1/c_a)\\
p_d                         &\sim beta(a_{0}, b_{0})\\
r^d_{a,k}                   &= r_{a_1,k}  \oplus  r_{a_2,k}  \oplus  \cdots \\
r_{d,k} | r_a, p_d          &\sim Gamma(r^d_{a,k}, p_d/(1 - p_d))\\
n_{d,k}                     &\sim Pois(r_{d,k}) \\
N_d                         &= \sum_{k=1}^K n_{d,k} \\
\theta_{1:K}                &\sim \frac{1}{\gamma_0} H \\
z_{d,n}                     &\sim Multi(r_{d,1}/\sum r_d, r_{d,2}/\sum r_d, r_{d,3}/\sum r_d, \cdots) \\
w_{d,n}                     &\sim \theta_{z_{d,n}} \\
\end{aligned}
\end{equation*}
where $\gamma_0 = \int d H$ is the total mass of measure $H$, and the parameters are given the appropriate priors. Here, $H$ is a $N$-dimensional Dirichlet distribution, and each $\theta$ is a topic that is a $N$-dimensional vector.

The difficult part of the inference for this model is the mixed part $\Gamma^d_a$ or $r^d_{a}$. Since $r^d_{a}= r_{a_1} \oplus r_{a_2} \oplus \cdots $ is the mixed value, it is hard to infer the posterior of $r_a$ through its likelihood. In order to resolve this issue, we firstly introduce the Additive Property of the negative binomial distribution,

\begin{theorem}
If $X_i$ follows a negative binomial distribution with parameters $r_i$ and $p$ and if the various $X_i$ are independent, then $\sum X_i$ follows a negative binomial distribution with parameters $\sum r_i$ and $p$.
\end{theorem}

In the model, we have
\begin{equation}
\begin{aligned}
r_{d,k} | \{r_a\}, p_d          &\sim Gamma(r^d_{a,k}, p_d/(1 - p_d))\\
n_{d,k}                     &\sim Pois(r_{d,k}) \\
\end{aligned}
\end{equation}
(in distribution) equal to
\begin{equation}
\begin{aligned}
n_{d,k}                     &\sim NB(r^d_{a,k}, p_d) \\
\end{aligned}
\end{equation}
and according to THEOREM 1, it is further (in distribution) equal to
\begin{equation}\label{nadk}
\begin{aligned}
n^a_{d,k}     &\sim NB\left(\frac{r_{a,k}}{A_d}, p_d\right) \\
n_{d,k}       &= \sum_a n^a_{d,k}    \\
\end{aligned}
\end{equation}
where $A_d$ is the number of authors in document $d$.

We have split $n_{d,k}$ the number of words assigned to topic $k$ in document $d$ into a number $A_d$ of independent variables $\{n^a_{d,k}\}$. Here, $n^a_{d,k}$ denotes the number of words assigned to topic $k$ from author $a$ in document $d$. From Eq.(\ref{nadk}), we can see that we have the likelihood part of the $r_{a}$, so we can update/inference the $r_{a}$ using $n^a_{d}$. Introducing the auxiliary variables $\{n^a_{d,k}\}$ helps us resolve the difficult inference problem brought by the mixed gamma process. Note that the independence between the elements of $\{n^a_{d,k}\}$ is very important, which facilitates us update each $n^a_{d,k}$ independently.

According to the relationship between the negative binomial distribution and the gamma-poisson distribution, for each $n^a_{d,k}$, we have:
\begin{equation}\label{nadkgp}
\begin{aligned}
&~~~~~~n^a_{d,k}     \sim NB(\frac{r_{a,k}}{A_d}, p_d) \\
&\Longrightarrow
r^a_{d,k}       \sim   Gamma(\frac{r_{a,k}}{A_d}, p_d/(1 - p_d)) ,~~
n^a_{d,k}       \sim   Pois(r^a_{d,k})
\end{aligned}
\end{equation}
We want to highlight that $r^a_{d,k}$ is different from $r^d_{a,k}$: $r^d_{a,k}$ is the mixed Gamma process of multiple author Gamma processes $\Gamma_a$ of Gamma process $\Gamma_d$ of document $d$ and $r^a_{d,k}$ is the interest of document $d$ on topic $k$ inherited from author $a$.

Due to the non-conjugacy of gamma distribution and negative binomial distribution, it is difficult to update $r_a$ with a gamma prior. In order to make the inference with only close-formed conditional distributions, we use the following results on the negative binomial process,
\begin{theorem}
\cite{quenouille1949relation,6636308} If $X$ follows a negative binomial distribution $X \sim NB(r, p)$ with parameters $r$ and $p$, then $X$ can also be generated from a compound poisson distribution as
\begin{equation}
\begin{aligned}
X = \sum_{t=1}^l u_t, t \overset{i.i.d}{\sim} Log(p),&~l \sim poiss\left(-rln(1-p)\right)
\end{aligned}
\end{equation}
where $Log()$ is a Logarithmic distribution. Furthermore, this poisson-logarithmic bivariate count distribution, $p(X, l)$, can be expressed as
\begin{equation}
\begin{aligned}
X \sim NB(r, p), ~l \sim CRT(X, r)
\end{aligned}
\end{equation}
\end{theorem}
where CRT denotes Chinese restaurant Table distribution.

With THEOREM 2, the Eq. (\ref{nadkgp}) is also equal to
\begin{equation}
\begin{aligned}
&~~~~~~n^a_{d,k}     \sim NB(\frac{r_{a,k}}{A_d}, p_d) \\
&\Longrightarrow
n^a_{d,k}       \sim   \sum_1^{l^a_{d,k}} log(p_d) ,~~
l^a_{d,k}       \sim    Pois(-\frac{r_{a,k}}{A_d} \cdot ln(1-p_d))
\\
&\Longrightarrow
l^a_{d,k}       \sim    CRT(n^a_{d,k}, \frac{r_{a,k}}{A_d}) ,~~
n^a_{d,k}       \sim    NB(\frac{r_{a,k}}{A_d}, p_d)
\end{aligned}
\end{equation}

Finally, we can update all $n^a_{d,k}$ by,
\begin{equation}\label{1}
\begin{aligned}
(n^{a_1}_{d,k_1},n^{a_1}_{d,k_2},\cdots,n^{a_A}_{d,K})       &\sim   Mult(n_d, \frac{r^{a_1}_{d,k_1} }{r}, \frac{r^{a_1}_{d,k_2}}{r}, \cdots, \frac{r^{a_A}_{d,K}}{r}) \\
r &= \sum_a \sum_k r^a_{d,k}
\end{aligned}
\end{equation}
and for each word $n$ in a document $d$, we can assign it to a topic $k$ and author $a$ by
\begin{equation}\label{1}
\begin{aligned}
p(z_{d,n} = k, i_{d,n} = a)    &\propto \frac{r^{a}_{d,k} }{r} \\
n_{d,k} &= \sum_n \delta(z_{d,n} = k ) \\
n_{a,k} &= \sum_d \sum_n \delta(z_{d,n} = k ~AND~ i_{d,n} = a) \\
\end{aligned}
\end{equation}

With these changes of variables, the original model is re-formulated as,
\begin{equation}
\begin{aligned}
\gamma_0                    &\sim Gamma(e_0, 1/f_0)\\
r_{0,k} | \gamma_0, c_0     &\sim Gamma(\gamma_0/K, 1/c_0)\\
p_d                         &\sim beta(a_{d,0}, b_{d,0})\\
r_{a,k} | r_0, c_a          &\sim Gamma(r_{0,k}, 1/c_a)
\\
r^d_{a,k}                   &= r_{a_1,k}  \oplus  r_{a_2,k}  \oplus  \cdots \\
r_{d,k} | r_a, p_d          &\sim Gamma(r^d_{a,k}, p_d/(1 - p_d))
\\
r^a_{d,k}                   &\sim Gamma(\frac{r_{a,k}}{A_d}, p_d/(1 - p_d)), ~~a \in A^d\\
z^a_{d,n}                   &\sim Discrete( \frac{r^{a}_{d,k} }{r}, \cdots )\\
n_{d,k}                     &=      \sum_n \delta(z_{d,n} = k ) \\
n_{a,k}                     &=      \sum_d \sum_n \delta(z_{d,n} = k ~AND~ i_{d,n} = a) \\
n^a_{d,k}                   &=      \sum_n \delta(z_{d,n} = k ~AND~ i_{d,n} = a) \\
N_d                         &=      \sum_n \sum_a z^a_{d,n}
\end{aligned}
\end{equation}

In the following, a Gibbs sampling algorithm is designed for the posterior inference and all the conditional distributions are listed.

\textbf{Sampling $z$}
\begin{equation}\label{zi}
\begin{aligned}
p(z_{d,n} = k, i_{d,n} = a | \cdots) \propto  \theta_{k,n}  \cdot r^a_{d,k}
\end{aligned}
\end{equation}

\textbf{Sampling $r^a_d$}
\begin{equation}\label{rad}
\begin{aligned}
p(r^a_{d,k} | \cdots) &\propto Gamma(\frac{r_{a,k}}{A_d} + n^a_{d,k}, p_d)
\end{aligned}
\end{equation}
where $n^a_{d,k}$ is the number of words in document $d$ with author $a$ and topic $k$.

\textbf{Sampling $l^a_d$}
\begin{equation}\label{lad}
\begin{aligned}
p(l^a_{d,k} | \cdots) &\propto CRT \left(n^a_{d,k}, \frac{r_{a,k}}{A_d} \right )\\
\end{aligned}
\end{equation}

\textbf{Sampling $p_d$}
\begin{equation}\label{pd}
\begin{aligned}
r^d_{a,k} &= r_{a_1,k}  \oplus  r_{a_2,k}  \oplus  \cdots
\\
p(p_d | \cdots) &\propto  Beta \left (a_{0} + \sum_k n_{d,k}, b_{0} + \sum_k r^d_{a,k} \right )
\\
p(r_{d,k} | \cdots) &\propto Gamma(r^d_{a,k} + n_{d,k}, p_d)
\end{aligned}
\end{equation}

\textbf{Sampling $r_a$}
\begin{equation}\label{ra}
\begin{aligned}
&~~~p(r_{a,k} | \cdots) \\
&\propto Gamma \left (r_{0,k} + \sum_{d ~\text{with}~ a} l^a_{d,k}, \frac{1}{c_a - \sum_{d ~\text{with}~ a} \frac{1}{A_d} \cdot \ln(1-p_d) } \right )
\end{aligned}
\end{equation}

\textbf{Sampling $l_a$}
\begin{equation}\label{la}
\begin{aligned}
p(l_{a,k} | \cdots) &\propto CRT \left(\sum_{d ~\text{with}~ a} l^a_{d,k}, r_{0,k} \right )\\
\end{aligned}
\end{equation}

\textbf{Sampling $r_{0,k}$}
\begin{equation}\label{r0k}
\begin{aligned}
p(r_{0,k} | \cdots)  &\propto  Gamma \left (\gamma_0/K + \sum_a l_{a,k}, \frac{1}{c_0 - \sum_a \ln(1-p_{a})} \right )
\end{aligned}
\end{equation}
where
\begin{equation}\label{pa}
\begin{aligned}
p_a &= \frac{-\sum_{d ~\text{with}~ a} \frac{1}{A_d} ln(1-p_d)}{c_a - \sum_{d ~\text{with}~ a} \frac{1}{A_d} ln(1-p_d)}
\end{aligned}
\end{equation}

\textbf{Sampling $l^\prime_{k}$}
\begin{equation}\label{lpk}
\begin{aligned}
p(l^\prime_{k} | \cdots) &\propto CRT \left( \sum_a l_{a,k}, \gamma_0/K \right )\\
\end{aligned}
\end{equation}

\textbf{Sampling $\gamma_0$}
\begin{equation}\label{g0}
\begin{aligned}
p(\gamma_0 | \cdots) &\propto Gamma \left(e_0 + \sum_k l^\prime_{k}, \frac{1}{f_0 - \ln(1-p^\prime)} \right )\\
\end{aligned}
\end{equation}
where
\begin{equation}\label{pp}
\begin{aligned}
p^\prime = \frac{-\sum_a ln(1-p_a)}{c_0 -\sum_a ln(1-p_a) }
\end{aligned}
\end{equation}

\textbf{Sampling $\theta_k$}
\begin{equation}\label{theta}
\begin{aligned}
p(\theta_k | \cdots) &\propto H(\theta_k) \prod_d \theta_{z_{d,n}=k, n}
\end{aligned}
\end{equation}

\begin{algorithm}[t]
\caption{Gibbs Sampler for IAT}
\KwIn{$D$, $A$, $N$, $AD$, $DN$}
\KwOut{$K_{real}$, \{$\theta$\}, \{$r_a$\}, \{$r_d$\}}
initialization\;
\While{$iter \le max_{iter}$}
{
    \For{$d=1; d \le D$}
    {
         \For{$n=1; n \le N_d$}
        {
             Update $z_{d,n}$ and $i_{d,n}$ by Eq. (\ref{zi});
        }
        \For{$a=1; a \le A_d$}
        {
            Update $r^a_{d,k}$ by Eq. (\ref{rad});
            Update $l^a_{d,k}$ by Eq. (\ref{lad});
        }
        Update $r_{d,k}$ and $p_d$ by Eq. (\ref{pd});
    }
    \For{$a=1; a \le A$}
    {
        Update $r_{a,k}$ by Eq. (\ref{ra});
        Update $l_{a,k}$ by Eq. (\ref{la});
    }
    Update $r_{0,k}$ by Eq. (\ref{r0k});
    Update $l^\prime_{k}$ by Eq. (\ref{lpk});
    Update $\gamma_0$ by Eq. (\ref{g0});
    Update $\theta$ by Eq. (\ref{theta});

    $iter++$;
}
Identify $K_{real}$;\\
Select the sample with largest likelihood and $K = K_{real}$;\\
return \{$\theta$\}, \{$r_a$\}, \{$r_d$\}\;
\label{ag:iat}
\end{algorithm}

We can see from these conditional distributions that all of them are closed-form which is very easy to updated and implemented. Note that the sampling of the CRT distribution can be found in \cite{6636308}. The whole procedure is summarized in Algorithm \ref{ag:iat}.

Note that after we obtain all the samples of the posterior $p(\theta, r_a, r_d, r_0, z^a_{d,n}, p_d, \gamma_0, n^a_{d,k}| N_d, AD, DN, e_0, f_0, c_0, c_a, a_0, b_0)$ of latent variables and remove the burn-in stage, we firstly identify the topic number with largest frequency as the $K_{real}$, and then find the sample with largest likelihood and $K=K_{real}$ from these samples. The output of Gibbs sampler are the latent variables $\theta$, $r_a$ and $r_d$ in this sample.

\section{Experiments}

\begin{table}[!t]
\caption{Statistics of Datasets}\label{datasets}
\centering
\begin{tabular}{c|c|c|c}
\hline
~~~~~~Datasets~~~~~               & ~~~~~~D~~~~~~   & ~~~~~~A~~~~~~   & ~~~~~~N~~~~~~\\
\hline
\emph{NIPS}                 & 1,740 & 2,037 & 13,649      \\
\hline
\emph{DBLP}                 & 28,569 & 28,702 & 11,771      \\
\hline
\end{tabular}
\end{table}

\begin{table}[!t]
\caption{Groups of Datasets DBLP}\label{dblpgroup}
\centering
\begin{tabular}{c|c|c|c|c}
\hline
       & D training   & ~D test~  & ~~~~~A~~~~~   & ~~~~~N~~~~~ \\
\hline
~\emph{group 1}~              & 1,072 & 319 & 1,115 & 3,783      \\
\hline
\emph{group 2}              & 1,071 & 316 & 1,094 & 3,782      \\
\hline
\emph{group 3}              & 1,075 & 305 & 1,071 & 3,788      \\
\hline
\emph{group 4}              & 1,076 & 339 & 1,104 & 3,823      \\
\hline
\emph{group 5}              & 1,079 & 310 & 1,111 & 3,841      \\
\hline
\end{tabular}
\end{table}

\begin{table}[!t]
\caption{Groups of Datasets NIPS}\label{nipsgroup}
\centering
\begin{tabular}{c|c|c|c|c}
\hline
       & D training   & ~~D test~~  & ~~~~A~~~~   & ~~~~N~~~~ \\
\hline
~~\emph{group 1}~~              & 1,503 & 237 & 2,037 & 5,110      \\
\hline
\emph{group 2}             & 1,495 & 245 & 2,037 & 5,110      \\
\hline
\emph{group 3}              & 1,511 & 229 & 2,037 & 5,110      \\
\hline
\end{tabular}
\end{table}

In this section, we evaluate the proposed infinite author topic model (IAT), and compare it with the finite author-topic model (ATM) on different datasets.

\subsection{Datasets}

Two public datasets used in this paper are:
\begin{itemize}
  \item \textbf{NIPS papers}\footnote{http://www.datalab.uci.edu/author-topic/NIPs.htm} This dataset contains papers from the NIPS conferences between 1987 and 1999. More description can be found in the \cite{atm2};
  \item \textbf{DBLP papers}\footnote{http://www.cs.uiuc.edu/~hbdeng/data/kdd2011.htm} The abstracts and authors of papers are extracted through DBLP interface from four areas: database, data mining, information retrieval and artificial intelligence. More description can be found in the \cite{Deng:2011:PTM:2020408.2020600}.
\end{itemize}

Some statistics of two datasets are shown in Table \ref{datasets}. For each dataset, we randomly select some documents as training data and test data. The Table \ref{nipsgroup} and Table \ref{dblpgroup} show the selection results on two datasets. The number of selected training and test documents are specialized in column \emph{D training} and column \emph{D test} in Table \ref{nipsgroup} and \ref{dblpgroup}. The requirements of selections is: the training and test documents must share some authors and some words. This requirement makes sure the learned topics and authors' interests can be used to predict the test documents.

\subsection{Evaluation Metrics}

In order to evaluate the performance of the proposed model, we calculate the perplexity of the test documents using the learned topics and author interests on these topics. Perplexity is widely used in language
modeling to assess the predictive power of a model \cite{atm2,blei2003latent}. It is a
measure of how surprising the words in the test documents are from the model's perspective. It can be computed as,
\begin{equation}\label{perp}
\begin{aligned}
Perplexity &= \exp \left (-\sum_d p(\mathbf{w_d}|a_d) \right )
\\
&= \exp \left(-\sum_d \sum_k p(\mathbf{w_d}|\theta_k) p(\theta_k | a_d) \right)
\end{aligned}
\end{equation}
where $a_d$ is the authors of test document $d$. The smaller the value of perplexity is, the better the predictive ability of a model has. Since we use the same test documents for different models, the normalization is not considered because it does not influence the model comparisons.

Another evaluation metric is the training data likelihood,
\begin{equation}\label{logl}
\begin{aligned}
logLikelihood = \sum_d \log p(w_d| \mathbf{\theta}, \mathbf{r_a}, \mathbf{r_d})
\end{aligned}
\end{equation}
This is a measure of the probability of the training document under the learned latent variables $\mathbf{\theta}$, $\mathbf{r_a}$ and $\mathbf{r_d}$. It can be understood as `how the model fits the training data'. The bigger the value of likelihood is, the better a model fits the training data.

\subsection{Results Analysis}

\begin{figure*}
\centering
\includegraphics[scale=0.5]{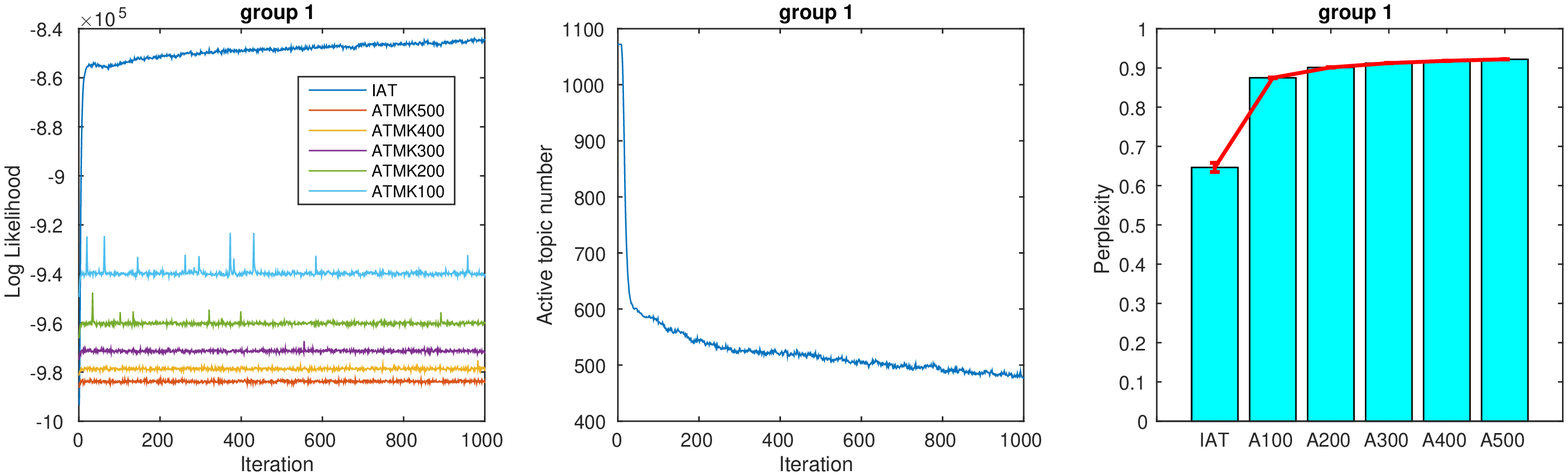}\\
\includegraphics[scale=0.5]{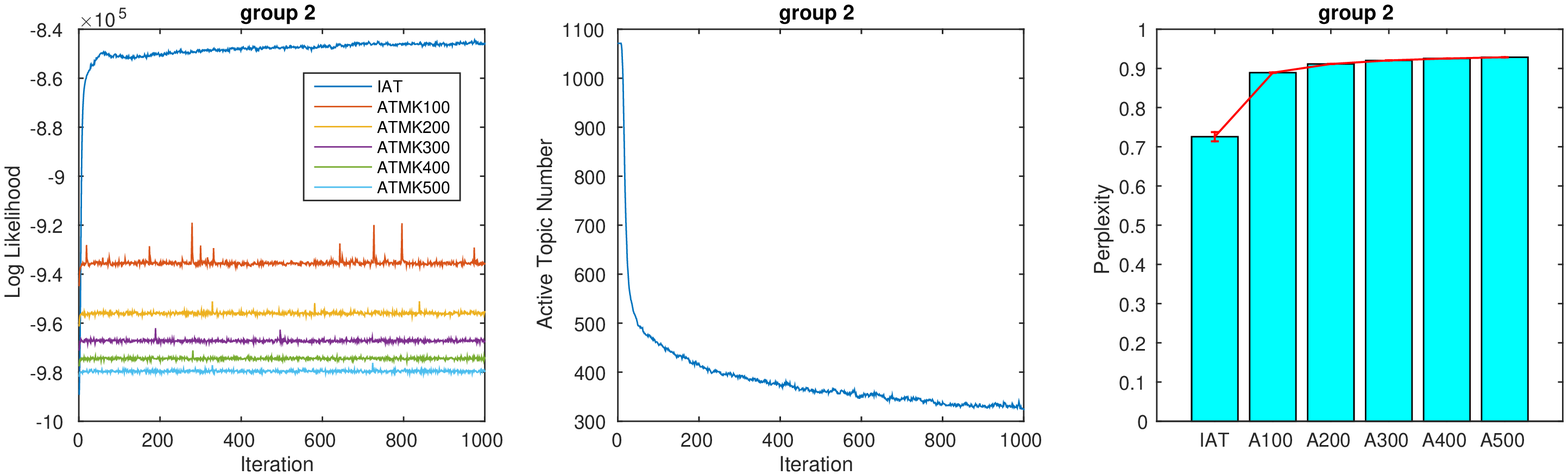}\\
\includegraphics[scale=0.5]{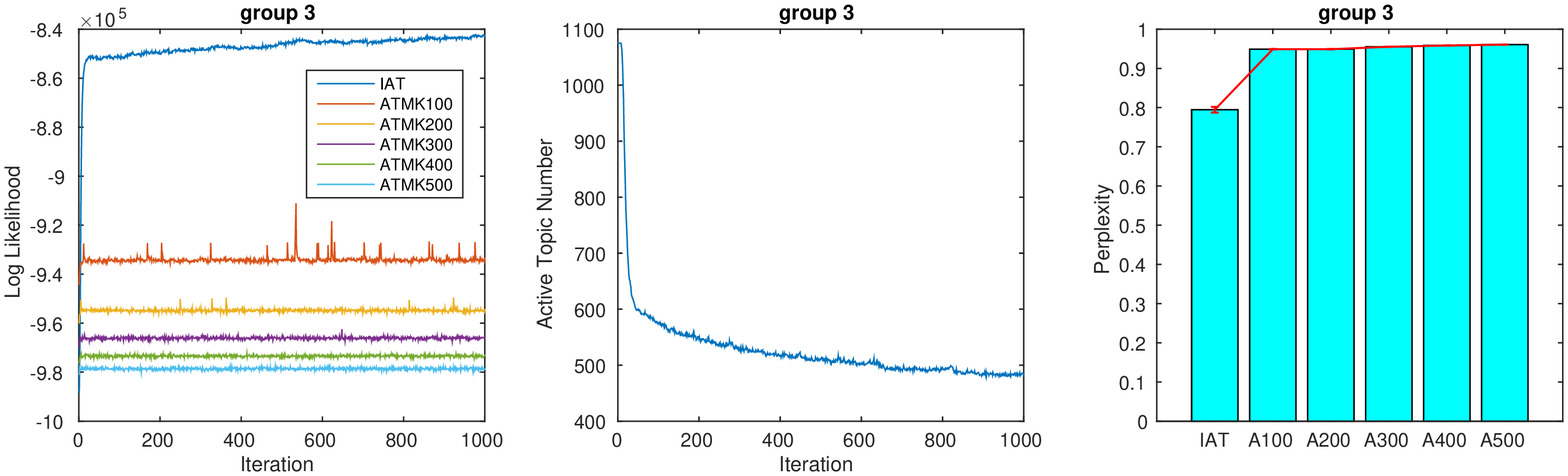}\\
\includegraphics[scale=0.5]{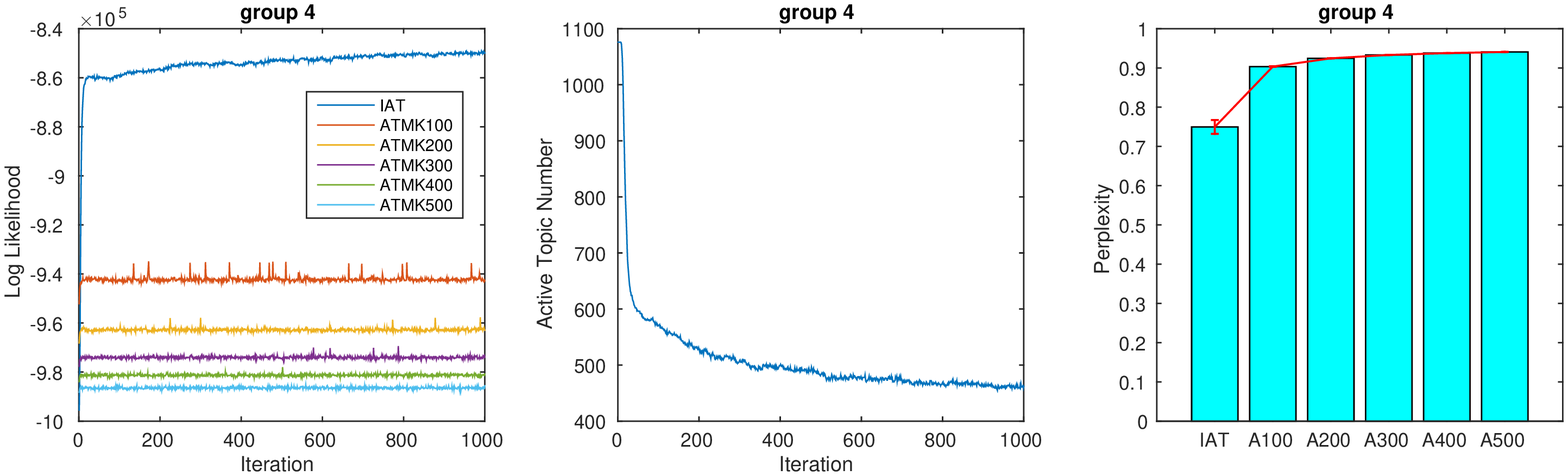}\\
\includegraphics[scale=0.5]{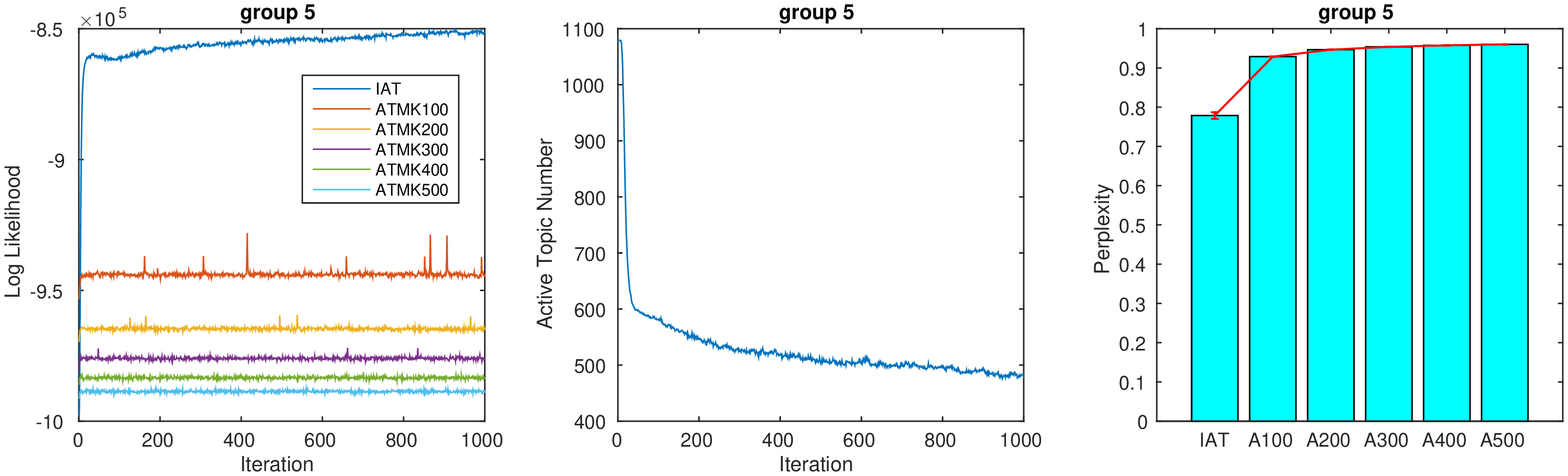}
\caption{Results from IAT and ATM on five groups of DBLP dataset. Each row denotes a group. In each row, the left subfigure shows the Log-likelihoods comparison between IAT and ATM with different (predefined) topic numbers: $K=100$, $K=200$, $K=300$, $K=400$, and $K=500$; The middle subfigure shows the change of active topic number of IAT during the iteration of Gibbs sampling; the right subfigure shows the perplexity comparison between IAT and ATMs. }
\label{fig:dblp}
\end{figure*}

\begin{figure*}
\centerline{\includegraphics[scale=0.75]{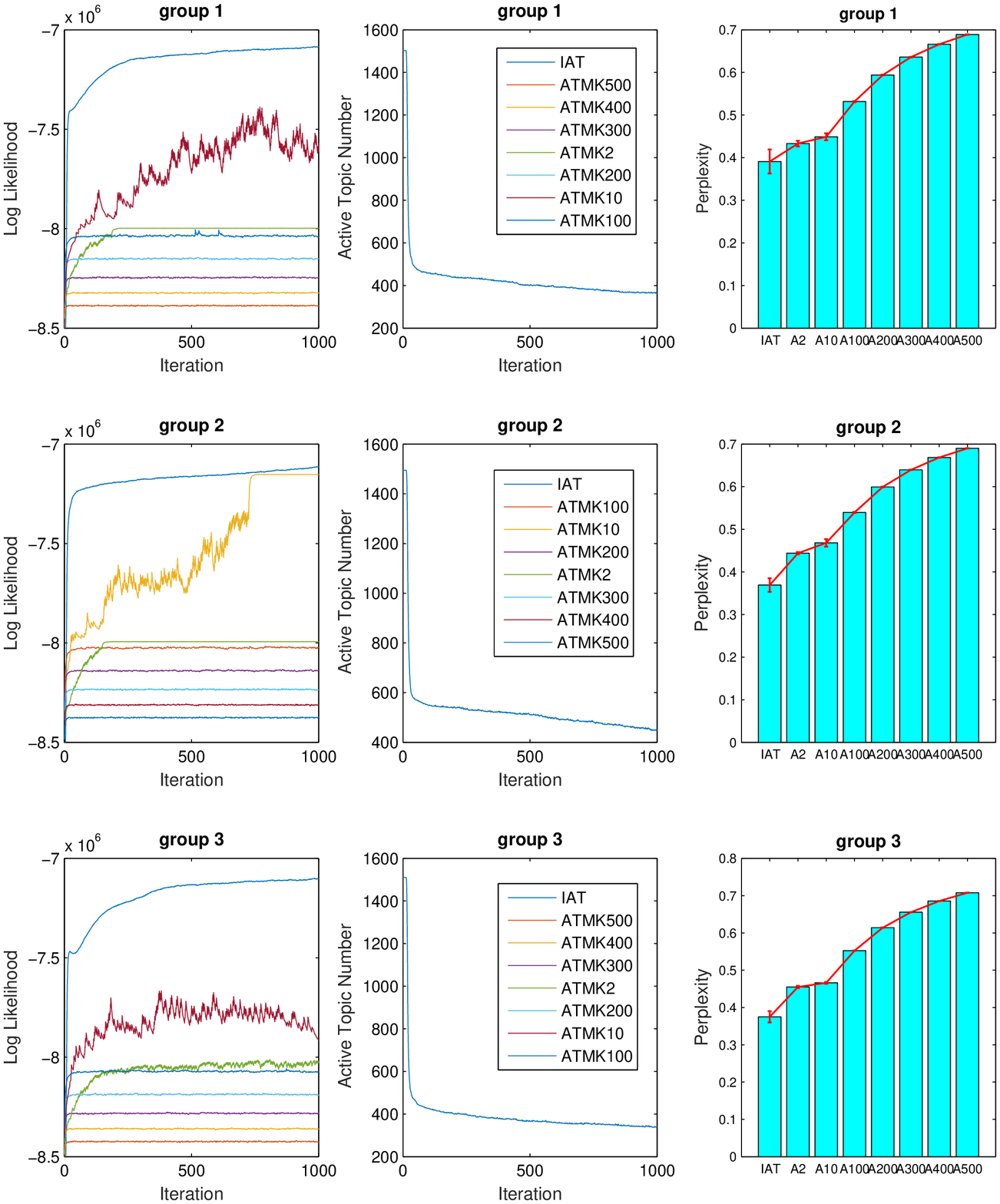}}
\caption{Results from IAT and ATM on three groups of NIPS dataset. Each row denotes a group. In each row, the left subfigure shows the Log-likelihoods comparison between IAT and ATM with different (predefined) topic numbers: $K=2$, $K=10$, $K=100$, $K=200$, $K=300$, $K=400$, and $K=500$; The middle subfigure shows the change of active topic number of IAT during the iteration of Gibbs sampling; the right subfigure shows the perplexity comparison between IAT and ATMs.}
\label{fig:nips}
\end{figure*}

For the DBLP dataset, the results are all shown in Fig. \ref{fig:dblp}. Each row of the Fig. \ref{fig:dblp} denotes a group of DBLP dataset corresponding to Table \ref{dblpgroup}. The left subfigures show the comparison on the data log-likelihood. Here, we adjust different active topic numbers for the ATM, including $K=100$, $K=200$, $K=300$, $K=400$ and $K=500$. From these subfigures, the proposed IAT model (The hyper-parameters are set as following by experiences for the rest of this section: $a_0=1$, $b_0=1$, $e_0=1$, $f_0=1$, $c_0=1$ and $c_a=1$) outperforms the ATM on different preset topic numbers. It means that IAT fits the training documents better than the ATM, and, more importantly, IAT does not depend the domain knowledge to predefine the active topic number, making the method widely applicable.

The middle subfigures in Fig. \ref{fig:dblp} indicate the changing of active topics during the iteration of the IAT (The number of active topics is set as the number of training documents at the initialization step of the model). These curves show that the number of active topics dramatically drops down at the burn-in stage of the sampling, and began to stabilize after about 200 iterations. Since the documents are different in content but similar in numbers amongst the groups, the learned topic number is differ slightly amongst each others. These numbers are: group 1: $K=519$; group 2: $K=332$; group 3: $K=493$; group 4: $K=465$; group 5: $K=504$.

In order to show the effectiveness of the proposed model, we also compare the performances of two models (IAT and ATM) on the test documents prediction using perplexity in Eq. (\ref{perp}). Since the training and test documents share some authors, we can compute the perplexity of the test documents according to the learned topics and authors' interests on them. At each step of iterations, the perplexity of test documents is computed using the latent variables, $\{\theta\}$, $\{r_a\}$ and $\{r_d\}$, at this iteration. The results are shown in right subfigures of Fig. \ref{fig:dblp}. In each subfigure, the first bar denotes the mean of perplexities of all iterations except the burn-in stage ($1\sim 200$ iterations) of the proposed model IAT and the others denote ATM with different (predefined) topic numbers. The standard deviations are also shown in the subfigures. The proposed model gets the best performance (smallest perplexity). The standard deviation of IAT is relatively bigger than ATM. The reason is because the number of active topics will change during the iteration but it will not change in ATM, so in theory, the random-walk space of Gibbs sampler of IAT can be larger than that of ATM. Even with this relatively larger standard deviation, the mean of perplexity of IAT is smaller than ATM.

For the NIPS dataset, the results are all shown in Fig. \ref{fig:nips}.
Same with the DBLP dataset, the log likelihoods of IAT and ATM with different predefined active topic numbers are shown in the left side of the Fig. \ref{fig:nips}. Unsurprisingly, the subfiguers in the middle column show the convergence of IAT (group 1: 367; group 2: 529; group 3: 354). Specially, we found that the log-likelihoods of ATM increases when topic number decreases. Therefore, we have compared with ATM with only two (the minimum number) topics as shown in the left subfigures in Fig. \ref{fig:nips}. It can be seen that the proposed IAT model also gets larger log likelihood and smaller perpetuity when compared with ATM except the case where ATM is set to have 10 topics in group 2. Even so, the ATM in group 2 with 10 topics has almost same performance with IAT on the Log-likelihood of training documents. Moreover, we can see that it takes 800 iterations to reach this stability for the ATM with 10 topics, but IAT only takes less than 50 iterations to reach the same stability.

\section{Conclusions and Further Study}

We have developed an infinite author topic model that can automatically learn completely the latent features of the author-document-keywords hierarchy, which include hidden topics, authors' interests on these topics and the number of topic from text corpora.
The stochastic processes are adopted instead of the fixed-dimensional probability distributions. The model uses a mixed author gamma process as the base measure of the document gamma process to capture the author-document mapping. We have demonstrated that the designed Gibbs sampling algorithm can be used to learn such infinite author topic model based on the various real-world datasets.

Other potential applications of this work include multi-label learning \cite{6471714}: The `authors' in the proposed model can be seen as labels, and the inference of the model can be seen as the training of the multi-label classifier. The learned topics can be seen as having infinite features space. This is our further study.

\section{Acknowledgments}

Research work reported in this paper was partly supported by the Australian Research Council (ARC) under discovery grant DP140101366 and the China Scholarship Council. This work was jointly supported by the National Science Foundation of China under grant no.61471232.

\bibliographystyle{abbrv}
\bibliography{sigproc}  

\end{document}